\pdfoutput=1

\documentclass[11pt]{article}

\usepackage{acl}
\usepackage{times}
\usepackage{latexsym}
\usepackage{circledsteps}
\usepackage{xspace}
\usepackage{amsmath}
\usepackage{amssymb}
\usepackage{multirow}
\usepackage{booktabs}       
\usepackage{graphicx}

\usepackage[T1]{fontenc}

\usepackage[utf8]{inputenc}

\usepackage{microtype}

\title{InheritSumm: A General, Versatile and Compact Summarizer \\by Distilling from GPT}

\author{Yichong Xu, Ruochen Xu, Dan Iter, Yang Liu, Shuohang Wang, Chenguang Zhu, Michael Zeng\\
Microsoft Cognitive Services Research Group\\
  \texttt{\{yicxu,ruox,iterdan,yaliu10,shuowa,chezhu,nzeng\}@microsoft.com} \\}

\newcommand{\modelname}[0]{\textsc{InheritSumm}\xspace}
\newcommand{\modelconsistent}[0]{\modelname-Consistent}
\newcommand{\modelbalanced}[0]{\modelname-Balanced}
\newcommand{\dataname}[0]{\textsc{GPTSumm}\xspace}

\begin{document}
	\maketitle
	\begin{abstract}
		While large models such as GPT-3 demonstrate exceptional performance in zeroshot and fewshot summarization tasks, their extensive serving and fine-tuning costs hinder their utilization in various applications. Conversely, previous studies have found that although automatic metrics tend to favor smaller fine-tuned models, the quality of the summaries they generate is inferior to that of larger models like GPT-3 when assessed by human evaluators. To address this issue, we propose \modelname, a versatile and compact summarization model derived from GPT-3.5 through distillation. 
		\modelname not only exhibits comparable zeroshot and fewshot summarization capabilities to GPT-3.5 but is also sufficiently compact for fine-tuning purposes. Experimental results demonstrate that \modelname achieves similar or superior performance to GPT-3.5 in zeroshot and fewshot settings. Furthermore, it outperforms the previously established best small models in both prefix-tuning and full-data fine-tuning scenarios.
	\end{abstract}
	
	\section{Introduction}
	Recently, the development of large language models (LLMs) like GPT-3 \cite{brown2020language} and PaLM \cite{chowdhery2022palm} has largely revolutionized the text summarization community, bringing a new paradigm in the the way summaries are generated. LLMs have demonstrated unparalleled ability to produce highly readable summaries while requiring little to no training data, overwhelmingly prefered by human annotators than those from smaller models \cite{goyal2022news}. Human evaluators note that GPT-3 generates summaries with superior readability and coherence, sometimes even more favored than human-generated summaries \cite{liang2022holistic}. One of the key advantages of LLMs is their capacity for zeroshot and fewshot learning \cite{brown2020language}, which enables them to adapt to new domains with ease. Therefore, LLMs are highly attractive for a wide range of summarization applications.

	
	Despite these remarkable achievements, training and deploying LLMs for summarization is computationally expensive. Deploying an LLM for inference is already impractical for many NLP practitioners, especially for summarization where long input length of the document and demonstrative examples are typical.
	Moreover, the prohibitive cost to finetune an LLM makes them hard to adapt to various custom domains, when abundant training data is available. This largely hurts the widespread adoption of LLMs for summarization.
	Thus, \citet{impossibletriangle} proposes the target of achieving all three aspects of the impossible triangle for NLP models: moderate model size, superior zeroshot / fewshot learning capability and superior supervised learning capability.

	In light of these challenges, we propose \modelname, a versatile summarization model with a smaller size, but with similar generalization capabilities. \modelname is trained using knowledge distillation from the GPT-3 model, by mimicing the GPT-generated summaries on general documents. To facilitate this, we curated the \dataname dataset, with over 7 million document-summary pairs. The documents are collected from general language corpora and GPT-3 generates the corresponding summary for them. Utilizing this dataset, we train a ZCode++ \cite{he2022z} model by following the GPT-generated summaries in both zeroshot and fewshot learning settings.
	
	One important limitation of fewshot learning for summarization is the input length limit: The in-context demonstrations eats up the input length and usually it is not feasible to include many in-context examples in the prompt. We propose to use \emph{succinct demonstrations} to alleviate this issue by shortening the demonstration documents, and therefore enabling more in-context examples.
	
	By training on the knowledge in \dataname and using succinct demonstrations, we show that \modelname is able to closely match, or sometimes even outperform GPT-3 in zeroshot and fewshot summarization. Therefore, our \modelname with only 398M parameters is able to achieve the impossible triangle \cite{impossibletriangle} for the summarization task. We show that \modelname has strong performance and most versatile summarization capablilty to date: it achieves best or close-to-best performance in all the settings including zeroshot/fewshot learning with prompts, fewshot learning with prefix tuning, as well as fully supervised learning.
	\begin{figure*}[tb!]
		\centering
		\includegraphics[width=0.85\linewidth]{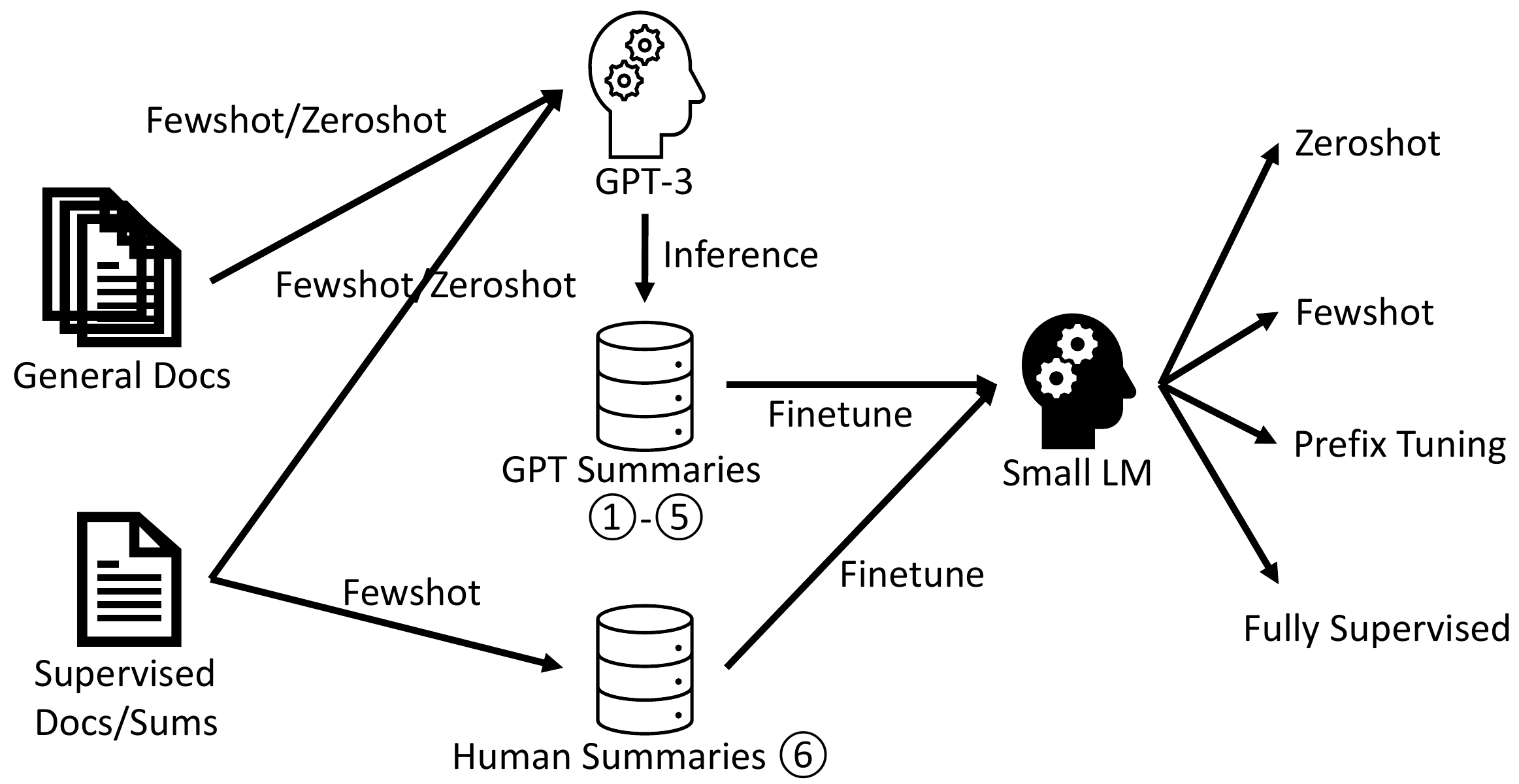}
		\caption{Overview of our method.}
		\label{fig:method_plot}
	\end{figure*}

	
	
	Our key contribution are three-fold. Firstly, we build the \dataname dataset with more than 4M paragraph-summary pairs by using querying GPT-3.5. To the best of our knowledge, \dataname is the largest corpus to date focusing on distilling from GPT-3.5, and is the first one focusing on summarization. Secondly, we build \modelname based on the \dataname corpus. We show that \modelname exhibits versatile capabilities across zeroshot, fewshot, and supervised settings, achieving the impossible triangle for summarization \citep{impossibletriangle}. Lastly, we propose new methods to include in-context examples for fewshot summarization that improves the performance.
	
	\section{Overview \& Problem Statement}

	In summarization, the goal is to summarize the input document $D$ into a short summary $Y$. For zeroshot and fewshot learning, we can add a prompt $P$ to the document to familiarize the model (parameterized with $\theta$) with the given task. A sequence-to-sequence model then takes the input $X=[P; D]$ \footnote{Our method also applies to the case where the prompt content appears after the document $D$.} and predicts the summary $\hat{Y}$. 
	
	In zeroshot learning, $P$ is a short description of the task, e.g., ``summarize the following document into a short paragraph''. For fewshot learning with in-context examples, the prompt consists of several descriptive examples along with an instruction, i.e., $P=[X_1;Y_1, ..., X_n; Y_n; I]$, where $X_i$ and $Y_i$ are illustrative examples, and $I$ is the task description. 
	For prefix tuning and supervised learning, $P$ is empty and $X=D$. We tune parameters $\theta$ to adapt the model to a given task.
	
	
	Figure \ref{fig:method_plot} describes our overall method. We distill summarization skills from GPT-3 by using it to generate summaries for documents from different domains, including general documents used for language modeling, as well as specialized documents from labeled datasets. The GPT-generated summaries and (relatively fewer) human generated summaries from labeled datasets forms our \dataname dataset. We then use \dataname to train a seq2seq model, and then adapt it to zeroshot/fewshot learning, prefix tuning and supervised learning summarization.
	
	In the following subsections, we first describe the method to build the \dataname dataset in Section \ref{sec:data_collection}, and then introduce the model training process in Section \ref{sec:model_training}. Finally we describe our method to adapt the pretrained \modelname model to zeroshot, fewshot, and supervised summarization.
	
	
	\section{Distillation Data Collection of \dataname\label{sec:data_collection}}
	We discuss the construction of \dataname as the data for distillation in this section.
	\begin{table}[t]
		\caption{Sources of Documents in \dataname.}
		\centering
		\label{tab:doc_collection}
		\scalebox{0.8}{\begin{tabular}{cccc}
			\toprule
			Datasets & Domain & \# Docs (\%)  & Length \\
			\midrule
			The Pile & General & 5.3M & 1,296 \\
			ArXiv & Academic & 203k & 6039\\
			CNN/DM & News & 287k & 781\\
			WikiHow & Instructions & 230k & 578 \\
			\bottomrule
		\end{tabular}}
	\end{table}
	
	\begin{table*}[t]
		\caption{Modes of generation in \dataname. ICD means in-context demonstrations. General means documents from the Pile corpus. Supervised corresponds to the labeled datasets in Table \ref{tab:doc_collection}. The quantity listed are after the filtering steps specified in Sec \ref{sec:data_collection}. 
  }
		\centering
		\label{tab:sum_generation}
		\begin{tabular}{ccccccc}
			\toprule
			Index & ICD Docs & ICD Summaries & ICD Num &  Input Docs & Target Summaries  & Quantity\\
			\midrule
			\Circled{1} & None & None & 0 & General &  GPT (zeroshot) & 0.5M \\
			\Circled{2} & General & GPT (from \Circled{1}) & 1  & General &  GPT (fewshot) & 2.6M\\
			\Circled{3} & Supervised & Supervised & 1 & General &  GPT (fewshot) & 2.2M\\
			\Circled{4} & None & None & 0 & Supervised  &  GPT (zeroshot) & 0.6M\\
			\Circled{5} & Supervised & Supervised & up to 4 & Supervised &  GPT (fewshot) & 0.5M\\
			\Circled{6} & Supervised & Supervised & up to 4 & Supervised &  Supervised & 0.6M\\
			
			\bottomrule
		\end{tabular}
	\end{table*}
	
	\subsection{Document Collection}
	To increase the generalization of downstream models, we collect a corpus of documents from various sources. We first include 3.1M documents from the Pile corpus \cite{pile} 
	We filter out non-English documents or the ones with too many non-character symbols \footnote{Namely, we remove documents where the percentage of non-English character symbols is larger than 70\%.}. We truncate each document to be within 4096 words and remove duplicates following \cite{smith2022using}. To include document from diverse domains and get closer to downstream tasks, we also include documents from arXiv\cite{cohan-etal-2018-discourse}, CNN/Daily Mail\cite{see-etal-2017-get} and WikiHow\cite{koupaee2018wikihow} datasets. Table \ref{tab:doc_collection} describes the composition of documents in detail.
	
	\subsection{Summary Generation} 
	We utilize the GPT-3.5 model, specifically the \texttt{text-davinci-002} variant, to generate summaries.  To adapt to downstream use cases, we apply different inputs, prompts, number of in-context demonstrations and zeroshot / fewshot settings to GPT to continue improving the quality of generated data, as shown in Table \ref{tab:sum_generation}. 
	
	Initially, we collect instructions from the PromptSource dataset \cite{bach2022promptsource} and filter out those that only produce a "subject line" rather than a complete summary. This process yields a final set of 24 instructions. 
	To generate examples for fewshot learning, we use these instructions to produce summaries for 500k documents in the Pile corpus (referred to as "General") in a zeroshot manner, resulting in data type \Circled{1}.
	After removing summaries with low quality (as detailed in section \ref{subsec:data_filter}), we use these summaries as in-context examples to generate summaries for an additional 2.6M documents in the Pile, which we denote as data type \Circled{2}. In addition to the zeroshot examples from GPT, we also leverage document-summary pairs from supervised datasets (Table \ref{tab:doc_collection}) as demonstrations to obtain data type \Circled{3}.
	
	For parts \Circled{4}\Circled{5}\Circled{6}, we use the supervised datasets as the input documents. In \Circled{4} and \Circled{5}, we utilize GPT to generate the summaries in either zeroshot or fewshot settings, using the supervised datasets as in-context demonstrations. Finally, in \Circled{6}, we employ the supervised datasets themselves to learn in a multitask learning fashion. We use a specific ``following'' instruction (detailed in Appendix \ref{app:prompts}) 
	to follow the in-context demonstrations for part \Circled{6} to enable the model to produce diverse outputs compared to \Circled{5}. Through this approach, our model can follow the given instructions to generate distinct summaries for the same documents.
	
	
	\subsection{Data Filtering} 
	\label{subsec:data_filter}
	To ensure the quality and relevance of the generated summaries, we implemented a filtering process comprising several steps. The first step involves retaining only those summaries with a "finish reason" of "stop." This indicates that the model has reached a natural stopping point, which serves as a reliable signal for the summary's quality.
	
	Subsequently, the text undergoes post-processing to eliminate any superfluous content. If the text contains a string that indicates another round of generation, such as consecutive newlines, the text is divided at the consecutive newlines, and only the first part is retained. 
	
	We then remove summaries whose word count is less than 10, greater than 512, or exceeds the length of the document. 
	
	Following this, the generated summary is assessed using the ROUGE metric to gauge its overlap with the original document. In this case, we treat the original document as the ground truth and the produced summary as the prediction. As a high-quality summary is expected to encompass a subset of the document's information, the recall score largely depends on the document's length and is not particularly informative for evaluating summary quality. Consequently, we rely solely on precision scores for filtering. Specifically, we retain the summary only if its ROUGE-1 precision with the document exceeds 0.6 and its ROUGE-2 precision falls within the range of 0.25 to 0.8. We establish an upper bound for ROUGE-2 precision because we observed that the GPT model occasionally copies sentences from the original document, which may include excessive, irrelevant details. 
	
	The filtering process is crucial for ensuring that the generated summary is of high quality and pertinent to the input document. Filtering is particularly important during the bootstrapping generation of data type \Circled{1}, where we filtered out 17\% of the produced summaries. The filtering rate is significantly lower for other data types. For example, less than 5\% of the summaries were filtered for data type \Circled{2}, where one-shot in-context demonstration was applied.
	
	\subsection{Succinct Demonstrations}
	\label{sec:succinct}
	Prior work \citep{NEURIPS2020_1457c0d6} has discovered that an increasing number of in-context examples can enhance the overall quality of generation. However, documents in the summarization task are typically lengthy, leading to a reduced usable input length for GPT and distilled models when including numerous in-context examples.
 To alleviate this problem, we propose to truncate the document and summary to include more examples. To do this, we first truncate all training documents and summaries in the supervised datasets to length $k$ ($k=256$ in our experiments). For every truncated document/summary, we add a suffix ``<omitted, $l$ words in total>'' to the text, where $l$ is the length of the original text. More specifically, for every document $D$ longer than $k$ words, we construct a succinct truncated document $D'=[d_1, d_2, ..., d_k, $<omitted, $l(D)$ words in total>$]$, where $d_j$ is the $j$-th word in $D$, and similarly for every summary $S$. We then add up to $M$ ($M=4$ in our experiments) succinct document-summary pairs as in-context examples for any given input document, and ask the model (either GPT or distilled models) to follow the in-context demonstrations.
	We apply succinct demonstrations to data \Circled{5} and \Circled{6} with ICDs from supervised datasets. 
	We do not apply succinct demonstrations to \Circled{2} and \Circled{3} since we observe that altering the number of in-context examples does not significantly impact the quality of generated summaries for general documents from the Pile corpus.
	%
	\section{Model Training \label{sec:model_training}}
	We proceed to train a seq2seq transformer language model ZCode++ \cite{he2022z} parameterized by $\theta$, utilizing the \dataname data. The objective is to minimize negative log-likelihood given the inputs with prompts $[P; X]$:
	\begin{equation}
		L(\theta) = \sum_{i=1}^{|Y|} \log \mathbb{P}_{\theta}(y_i|P, X, y_1,..., y_{i-1})\label{eqn:loss}
	\end{equation}
	where $\mathbb{P}_{\theta}$ is the probability distribution given by model with parameters $\theta$. We train the model in a multi-task manner, where we mix all the data from \Circled{1} to \Circled{6} and sample from the data pool in every minibatch. To better balance the model towards general applications, we adjust the sampling ratio between each task to up-sample smaller data modes and down-sample larger ones. Analogous to the data generation in \dataname, we also truncate the in-context examples in data \Circled{5} and \Circled{6} as described in Section \ref{sec:succinct}.
	
	\noindent \textbf{In-Context Demonstrations in Distillation.} To improve the GPT generation quality, most of the data in \dataname are generated in a fewshot manner. A natural way to train the downstream model is to keep the input same as GPT. We call the corresponding model \modelconsistent. 
	However, the resulting model might not be good at zeroshot summarization since most data are in fewshot format. To obtain a model good at both zeroshot and fewshot summarization, we propose to randomly convert some of the fewshot examples to zeroshot examples.
	More specifically, we train another model where we randomly include 0 to 4 in-context examples by excluding examples in the prompts. We use zeroshot learning for data \Circled{1} and \Circled{4}, 0 or 1 examples for \Circled{2} and \Circled{3}, and 0 to 4 examples for \Circled{5}. We call the corresponding model \modelbalanced. Note that data \Circled{6} does not involve GPT generation, and we always include 1 to 4 in-context examples in our training.
	
	\subsection{Adapting to Application Scenarios}
	After training the \modelname model on \dataname,
	we adapt the \modelname to three different summarization settings.
	
	\noindent\textbf{Zeroshot and fewshot learning with prompts.} The ability to adapt to different tasks with no or few examples is one of the most important characteristics of large models like GPT. We test the pretrained \modelname model in exactly the same way as GPT. For zeroshot learning, we randomly sample one instruction from PromptSource \cite{bach2022promptsource}. For fewshot learning, we include up to 4 examples (decided by the input document length) with succinct demonstrations from the corresponding training set. 
	
	\noindent\textbf{Fewshot learning with prefix tuning.} We follow the setting in \citet{chen2021meta} and \citet{chen2022unisumm} to prefix-tune our model. For every task $t$, we add $K$ task-specific prefix vectors at every layer of the encoder and decoder for \modelname, parameterized by $\theta^t$. Then we freeze all other parameters in \modelname and tune $\theta^t$ by minimizing (\ref{eqn:loss}).
	
	\noindent\textbf{Fully Supervised Learning.} In this setting we use all the data in downstream datasets to perform fully supervised finetuning. All the model parameters are finetuned by minimizing loss (\ref{eqn:loss}), without any prompts or instructions.
	
	\section{Experiments}
	We detail the experiment results in this section. We first specify the implementation details and hyperparameters in Sec. \ref{sec:implementation}, and then introduce the main experiment results in Sec. \ref{sec:expr_results}.
	\subsection{Implementation Details \label{sec:implementation}}
	\begin{table}[htb!]
		\centering
		\scalebox{0.85}{
			\begin{tabular}{@{}l|c@{}}  
				\toprule  
				Hyper-parameter & Value \\  
				\midrule  
				Warmup Steps & 10,000 \\  
				Learning Rates & 2e-5 \\  
				Batch Size & 144 (base), 120 (large) \\  
				Local Attention Window & 256 \\  
				Global Layers & 6/11 (base), 11/23 (large) \\  
				Weight Decay & 0.01 \\  
				Training Steps & 300k \\  
				Learning Rate Decay & Linear \\  
				Adam $\epsilon$ & 1e-6 \\  
				Adam $\beta_1$ & 0.9 \\  
				Adam $\beta_2$ & 0.999 \\  
				Gradient Clipping & 1.0 \\  
				Beam search size & 5 \\  
				\bottomrule  
			\end{tabular}  
			
		}
		\caption{
			Hyper-parameters for Training \modelname. 
		}
		\label{tab:hyperparam}
	\end{table}
	
	We use the ZCode++ model \cite{he2022z} as the language model backbone for training \modelname. We choose ZCode++ over other pretrained models since it achieves state-of-art performance on summarization tasks after finetuning. We experimented with both the pretrained Z-Code++ base and Z-Code++ large model.
	We train the model on the \dataname corpus with seq2seq loss (\ref{eqn:loss}) for 300K steps, with a maximum input length of 3072. We summarize the hyperparameters in Table \ref{tab:hyperparam}. To reduce the memory usage we follow ZCode++ to use Fusion-in-Encoder. We use two layers as the global attention layer and local layers have a window size of 256.
	
	\noindent\textbf{Test settings.} We test \modelname in 4 different settings: i) zeroshot learning with random instructions from PromptSource \cite{bach2022promptsource}, ii) 4-shot learning with instructions, iii) 10-shot learning with prefix tuning, and iv) fully supervised learning. i) and ii) are prompt-based settings typically employed by large models like GPT, whereas iii) and iv) are more traditional settings for smaller models.
	
	\noindent \textbf{Baselines.} We compare with GPT \texttt{text-davinci-002},
 the original ZCode++, BART \cite{lewis2019bart} and UniSumm \cite{chen2022unisumm}. Due to hardware and compute limits, we do not compare with GPT in prefix tuning and fully supervised settings.
	For fewshot learning with prefix tuning, We follow \citet{li2021prefix} to tune prefix embeddings at every encoder and decoder layer. For local layers in Fusion-in-Encoder of ZCode++ and \modelname, one set of prefix embeddings are inserted at every local window for every local layer. 
	\begin{table}[t]
		\caption{Statistics of testing datasets. Avg. D/S Length is the average number of GPT tokens for document/summary for the corresponding test set.}
		\centering
		\label{tab:test_datasets}
		\begin{tabular}{ccc}
			\toprule
			Datasets & Domain & Avg. D/S Length \\
			\midrule
			MultiNews & News &  1,979 / 275\\
			XWiki & Wikipedia &  971 / 85 \\
			SAMSum & Dialogue & 136 / 24\\
			Reddit-TIFU  & Forum & 496 / 29 \\
			BigPatent & Legal & 2,853 / 119 \\
			\bottomrule
		\end{tabular}
	\end{table}
	
	\noindent\textbf{Testing Datasets.} We test on 5 summarization datasets on diverse domains (a summary in Table \ref{tab:test_datasets}):\\
	\textbf{MultiNews} \cite{fabbri-etal-2019-multi} is a multi-document news summarization dataset with news from different sources.\\
	\textbf{XWiki} \cite{perez-beltrachini-lapata-2021-models} is a cross-lingual summarization dataset focusing on Wikipedia articles. We use the English data with paired documents and summaries. \\
	\textbf{SAMSum} \cite{gliwa-etal-2019-samsum} is a dialogue summarization with chit-chat dialogues in online chatting styles like Messenger and WhatsApp. Both the dialogue and summary are human-written by expert linguists. \\
	\textbf{Reddit-TIFU} \cite{kim-etal-2019-abstractive} is another dialogue summarization dataset focusing on the online forum Reddit. The language style on Reddit is significantly different from news articles.\\
	\textbf{BigPatent} \cite{sharma-etal-2019-bigpatent} is a legal document summarization dataset with US patent documents and human-written abstracts. Documents come from 9 different domains.  \\
	
	Some of the datasets have an extensive test set that takes a long time to test GPT on. Also, the GPT API has a content filter that rejects some of the input documents. Therefore, we randomly sample 500 documents from the test sets, and choose those that pass GPT's content filter to compare the baselines. 

	\subsection{Experiment Results \label{sec:expr_results}}
	\begin{table*}[t]
		\caption{Main Experiment Results. For simplicity, we only include ROUGE-2 scores in the table. IS stands for \modelname, B stands for balanced and C stands for consistent. The largest and second-largest number in each row are in \textbf{bold} and \textit{italic} respectively. The ``Mean'' section at the bottom of the table is the mean performance over 4 different settings.}
		\centering
		\label{tab:main_results}
		\scalebox{0.8}{\begin{tabular}{c|c|ccccccccc}
				\toprule
				Task & Datasets & Z-Base & Z-Large & GPT-3.5 & BART & UniSumm & IS-Base-B & IS-Base-C & IS-Large-B & IS-Large-C\\
				\multirow{6}{*}{\rotatebox[origin=c]{90}{Zeroshot}}
 & Samsum & 0.78 & 0.01 & 10.85 & 8.05 & 4.58 & \textbf{17.82} & 16.13 & 15.05 & \textit{16.73} \\
 & Xwiki & 1.41 & 0.12 & 6.97 & 5.47 & \textbf{9.31} & 8.35 & 8.02 & 8.50 & \textit{8.59} \\
 & Reddit Tifu & 0.09 & 0.00 & 4.02 & 3.03 & 4.41 & 5.75 & 4.94 & \textbf{5.97} & \textit{5.93} \\
 & Bigpatent & 1.18 & 0.01 & 10.09 & 9.08 & 10.83 & \textit{12.36} & 12.29 & \textbf{12.44} & 11.93 \\
 & MultiNews & 1.54 & 0.07 & 8.23 & \textbf{10.28} & 3.47 & \textit{8.65} & 8.31 & 7.01 & 7.72 \\
 & Avg & 1.00 & 0.04 & 8.03 & 7.18 & 6.52 & \textbf{10.59} & 9.94 & 9.80 & \textit{10.18} \\
				\midrule
				\multirow{6}{*}{\rotatebox[origin=c]{90}{Fewshot(instruct)}}   & Samsum & 0.00 & 0.05 & 18.67 & 1.08 & 0.60 & \textit{18.90} & \textbf{18.93} & 18.71 & 17.59 \\
 & Xwiki & 0.98 & 0.10 & \textbf{11.83} & 1.53 & 3.78 & 9.55 & 9.84 & 9.58 & \textit{9.99} \\
 & Reddit Tifu & 0.03 & 0.03 & 6.68 & 0.62 & 0.91 & 7.28 & 7.14 & \textbf{8.51} & \textit{7.93} \\
 & Bigpatent & 1.22 & 0.01 & 12.85 & 3.55 & 4.45 & 9.98 & 12.71 & \textbf{13.18} & \textit{12.88} \\
 & MultiNews & 0.92 & 0.07 & \textbf{11.32} & 1.92 & 1.33 & 10.21 & 9.76 & \textit{10.77} & 10.04 \\
 & Avg & 0.63 & 0.05 & \textbf{12.27} & 1.74 & 2.21 & 11.19 & 11.67 & \textit{12.15} & 11.68 \\
				\midrule
				\multirow{6}{*}{\rotatebox[origin=c]{90}{Fewshot(Prefix)}}  & Samsum & 14.25 & 15.79 & N/A & 9.88 & 11.37 & \textit{20.97} & \textbf{20.99} & 19.49 & 19.89 \\
 & Xwiki & 10.47 & 12.07 & N/A & 11.08 & 8.29 & 11.91 & 12.07 & \textit{12.20} & \textbf{12.46} \\
 & Reddit Tifu & 3.92 & 3.94 & N/A & 2.78 & 6.19 & 6.37 & \textit{6.54} & \textbf{6.92} & 5.26 \\
 & Bigpatent & 7.13 & 5.86 & N/A & 6.99 & \textbf{13.12} & 12.23 & 12.65 & \textit{12.84} & 12.23 \\
 & MultiNews & 4.88 & 8.92 & N/A & 11.63 & 10.84 & \textit{11.69} & \textbf{12.06} & 10.91 & 11.33 \\
 & Avg & 8.13 & 9.32 & N/A & 8.47 & 9.96 & \textit{12.63} & \textbf{12.86} & 12.47 & 12.23 \\
				\midrule
				\multirow{6}{*}{\rotatebox[origin=c]{90}{Supervised}}  & Samsum & 27.49 & 27.91 & N/A & 29.26 & 22.36 & \textbf{30.12} & \textit{29.87} & 28.52 & 28.60 \\
 & Xwiki & \textit{21.93} & 21.77 & N/A & 20.21 & 18.05 & 21.70 & 21.74 & \textbf{22.48} & 20.63 \\
 & Reddit Tifu & 10.66 & 10.37 & N/A & 11.33 & 8.42 & \textit{11.57} & \textbf{11.74} & 10.25 & 10.31 \\
 & Bigpatent & 17.94 & 12.64 & N/A & 17.88 & 17.38 & 17.99 & 18.03 & \textit{21.67} & \textbf{23.11} \\
 & MultiNews & 17.66 & 19.11 & N/A & 17.87 & 18.69 & 19.83 & \textit{20.38} & 19.03 & \textbf{20.58} \\
 & Avg & 19.14 & 18.36 & N/A & 19.31 & 16.98 & 20.24 & 20.35 & \textit{20.39} & \textbf{20.65} \\
\midrule
  \multirow{6}{*}{\rotatebox[origin=c]{90}{Mean}} & Samsum & 10.63 & 10.94 & N/A & 12.07 & 9.73 & \textbf{21.95} & \textit{21.48} & 20.44 & 20.70 \\
 & Xwiki & 8.70 & 8.52 & N/A & 9.57 & 9.86 & 12.88 & \textit{12.92} & \textbf{13.19} & 12.92 \\
 & Reddit Tifu & 3.67 & 3.58 & N/A & 4.44 & 4.98 & \textit{7.74} & 7.59 & \textbf{7.91} & 7.36 \\
 & Bigpatent & 6.87 & 4.63 & N/A & 9.38 & 11.45 & 13.14 & 13.92 & \textit{15.03} & \textbf{15.04} \\
 & MultiNews & 6.25 & 7.04 & N/A & 10.43 & 8.58 & \textit{12.59} & \textbf{12.63} & 11.93 & 12.42 \\
 & Avg & 7.23 & 6.94 & N/A & 9.18 & 8.92 & 13.66 & \textbf{13.71} & \textit{13.70} & 13.69 \\

				\bottomrule
		\end{tabular}}
	\end{table*} 
	
	We summarize our main results for the four settings in Table \ref{tab:main_results}. For simplicity and space reasons, we include ROUGE-2 scores in our experiment results. All the experiment results are from our own runs.
	
	\noindent\textbf{Zeroshot learning.} \modelname models demonstrated better performance than all the baselines in 3 out of 5 test datasets, as well as on the average over the 5 datasets. \modelname's performance is inferior to BART or UniSumm on MultiNews and Xwiki respectively, possibly because the summary on these two datasets are longer than the other datasets (this is also true for GPT-3.5 model). \modelname gets higher performance than the teacher GPT-3.5 model. 
 This is probably because \modelname is specialized in summarization, while GPT-3.5 might fail to follow the instructions to summarize the input document.\\
 Among the four variants of \modelname, the base \modelbalanced achieves the highest ROUGE score. The models trained in the balanced way receive more zeroshot examples in its training process, which probably makes them better at zeroshot learning. However, this is not true for \modelname-Large models, where the balanced model is slightly behind the consistent model. This might be because large models are more capable when generalizing across different settings, and the data composition (whether balanced or consistent) is not very important for large models.
	
	\noindent\textbf{Fewshot learning with instructions.} GPT-3.5 achieves the best average ROUGE score of 12.27 in this setting, whereas our \modelname models are only behind GPT-3.5 by a small gap. Among the four variants, \modelbalanced-Large achieves the best score of 12.15, slightly behind GPT-3.5. \modelbalanced-Large also beats GPT-3.5 on 2 of the test datasets. Large models are generally better in fewshot learning than base models. The performance between models trained with balanced or consistent data is comparable, likely because both models receive large quantities of fewshot data in their training.
	
	\noindent\textbf{Fewshot learning with prefix tuning.} \modelname generally achieves the best performance in this setting, only loses to UniSumm on Bigpatent by a small margin. \modelconsistent-base is the best in the average performance for the prefix tuning setting. The prefix tuning results of \modelname are also significantly better than the original ZCode++ models, suggesting the effectiveness of our distillation training.
	
	\noindent\textbf{Fully supervised learning.} Lastly, \modelname outperforms all the bselines in the fully supervised learning setting as well. \modelname outperforms the original ZCode++ model, showing the transfer ability of our distillation training. Among the four variant of \modelname, \modelconsistent-Large gives the best performance. This is likely because large models are more powerful with fully supervised data, and consistent data training is better for the transfer of knowledge.

 For average over the 4 settings, \modelname strongly outperforms all the baselines on the aggregate score over 4 settings, showing that \modelname is the most versatile model across different training scenarios. The average performance of the 4 variants is quite close.
	
	\subsubsection{Analysis}
	
	\begin{table}[htb]
		\caption{Performance of single-dataset training. R-2 stands for ROUGE-2 scores. All scores are averaging over 5 test sets.}
		\centering
		\label{tab:ablation_datasets}
		\begin{tabular}{ccc}
			\toprule
			Datasets & R-2(zeroshot) & R-2 (fewshot) \\
			\midrule
			\Circled{1} + \Circled{2} & 9.82 & 10.69 \\
			\Circled{3} & 10.02 & 10.04 \\
			\Circled{4} & 10.12 & 3.06\\
			\Circled{5} & 8.80 & 9.95 \\
			\Circled{6} & 6.53 & 6.17\\		
                \midrule
                All & 10.59 & 11.19 \\
			\bottomrule
		\end{tabular}
	\end{table}	
	\noindent\textbf{Effect of different training datasets.} One natural question is about the effect of each part of \dataname in \modelname's performance. While it is not possible to test every combination of \Circled{1}-\Circled{6}, we follow the FLAN paper \cite{wei2021finetuned} to test the performance of \modelname under individual parts. In Table \ref{tab:ablation_datasets}, we train a base \modelname in the balanced setting with the same hyperparameters on \Circled{1} + \Circled{2}, \Circled{3}, \Circled{4}, \Circled{5}, \Circled{6}\footnote{We combine \Circled{1} and \Circled{2} because they are quite similar in style and focus on the same set of input documents.} respectively. The results show that all the GPT-generated data (\Circled{1} - \Circled{5}) gives better zeroshot/fewshot performance than supervised datasets (\Circled{6}), except for \Circled{4} on fewshot: this is expected because \Circled{4} contains only zeroshot training data. As all parts of data can help boost the zeroshot/fewshot performance, we include all of them as our training data.
	\begin{figure}
	    \centering
	    \includegraphics[width=1.0\linewidth]{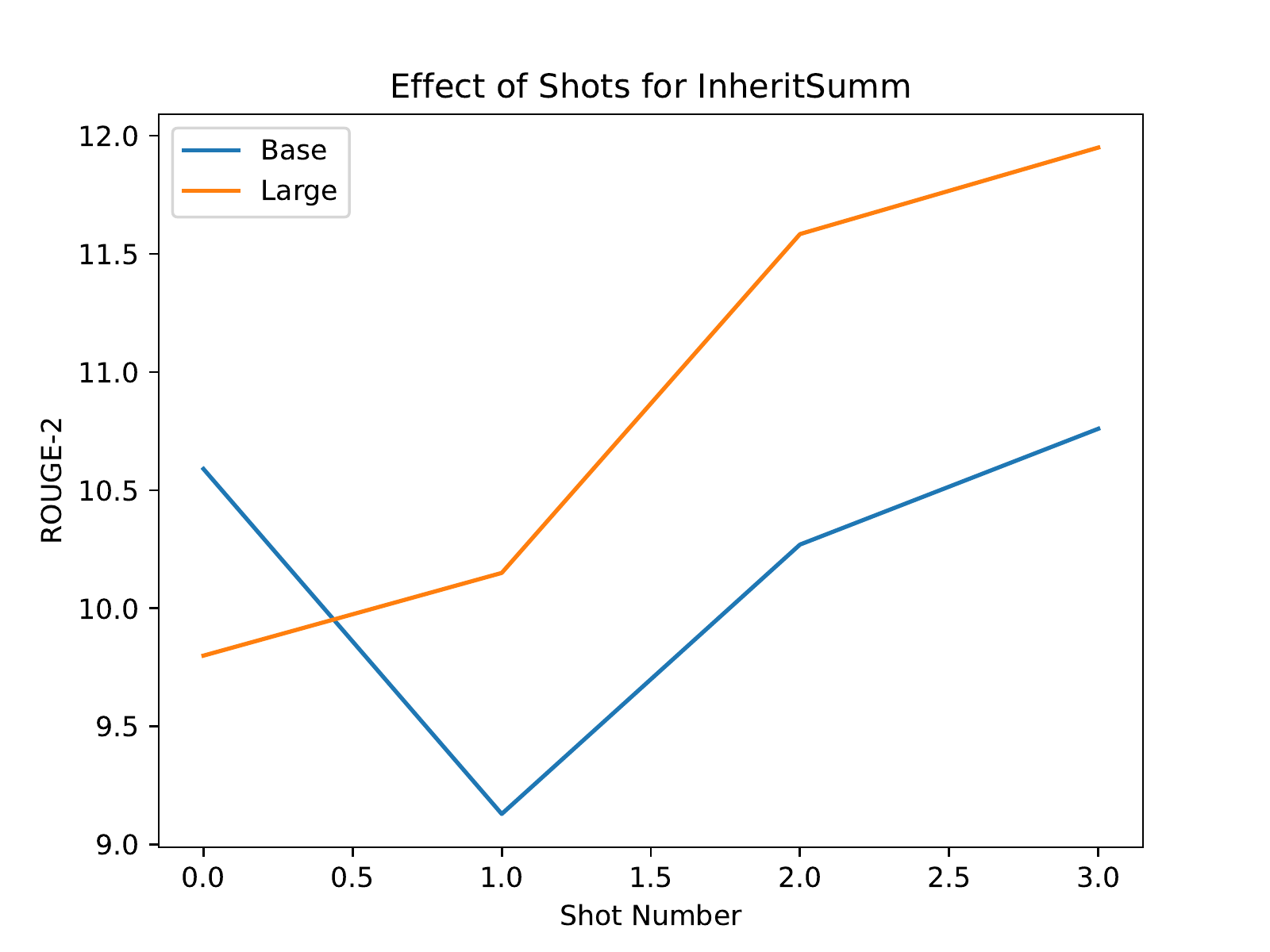}
	    \caption{Effect of number of shots for \modelname base and large model.}
	    \label{fig:plot_shots}
	\end{figure}
	\noindent\textbf{Effect of succinct demonstrations.} In order to test the effect of succinct demonstrations, we test \modelbalanced's performance with different number of shots. In Figure \ref{fig:plot_shots}, we plot the performance of base and large \modelbalanced from 0-shot to 4-shots. For both models, the performance improves from 1-shot to 4-shots. For the large model, the performance also improve when we go from zero-shot to 1-shot, but this is not the case for base model. This shows that using the traditional one-shot method may even hurt the performance, possibly due to model capacity reasons. Our succinct prompt method can always incorporate more in-context examples and improve the model's performance.

	\section{Related Works}
\noindent\textbf{Text Summarization} has been extensively explored by the community \cite{nenkova2012survey}. 
Previous works mostly focus on one or two settings of zeroshot/fewshot/supervised learning only. For example, BART \cite{lewis2019bart} and ZCode++ \cite{he2022z} focuses on supervised summarization. PEGASUS \cite{zhang2020pegasus} proposes pretraining methods for unsupervised summarization.  UniSumm \cite{chen2022unisumm} explores fewshot summarization with BART-Large. \cite{goyal2022news} explores zeroshot and fewshot summarization for GPT-3. To the best of our knowledge, we are the first paper to explore the generalization over zeroshot, fewshot, and supervised summarization.

\noindent\textbf{Model Distillation from GPT} There has been several works distilling knowledge from the GPT series of models. \citet{wang2022self} finetunes LLaMA \cite{touvron2023llama} with 52K instruction-following data using the \texttt{text-davinci-003} variant of GPT-3.5. It shows that the resulting Alpaca model behaves similarly to GPT-3.5 on instruction-following evaluation suite. \citet{peng2023instruction} further improves the performance by using instructions from the GPT-4 model. However, all these works focuses on the general setting with general user instructions. To the best of our knowledge, we are the first work on distillation from GPT-3/4 that focuses on a particular task. Our focus on summarization makes us able to use smaller models while not losing too much performance.

	\section{Conclusion \& Future Works}
We propose \modelname by distilling knowledge from GPT-3.5 using its summary on a broad range of documents. Base model of \modelname with only 400M parameters exhibits versatile capability on zeroshot, fewshot, and fully supervised summarization, surpassing performance of previous small models and beats or rivals GPT-3.5 in the overall performance. 

 For future works, it would be interesting to investigate the performance when distilling from other variants of GPT, like \textsc{text-davinci-003} or GPT-4. Another interesting direction is controllable summarization - by using proper instructions, \modelname can be further trained to generate customized summarizations with style or length constraints.
	
	\bibliography{custom,anthology}

\begin{thebibliography}{28}
\expandafter\ifx\csname natexlab\endcsname\relax\def\natexlab#1{#1}\fi

\bibitem[{Bach et~al.(2022)Bach, Sanh, Yong, Webson, Raffel, Nayak, Sharma,
  Kim, Bari, Fevry, Alyafeai, Dey, Santilli, Sun, Ben-David, Xu, Chhablani,
  Wang, Fries, Al-shaibani, Sharma, Thakker, Almubarak, Tang, Tang, Jiang, and
  Rush}]{bach2022promptsource}
Stephen~H. Bach, Victor Sanh, Zheng-Xin Yong, Albert Webson, Colin Raffel,
  Nihal~V. Nayak, Abheesht Sharma, Taewoon Kim, M~Saiful Bari, Thibault Fevry,
  Zaid Alyafeai, Manan Dey, Andrea Santilli, Zhiqing Sun, Srulik Ben-David,
  Canwen Xu, Gunjan Chhablani, Han Wang, Jason~Alan Fries, Maged~S.
  Al-shaibani, Shanya Sharma, Urmish Thakker, Khalid Almubarak, Xiangru Tang,
  Xiangru Tang, Mike Tian-Jian Jiang, and Alexander~M. Rush. 2022.
\newblock \href {http://arxiv.org/abs/2202.01279} {Promptsource: An integrated
  development environment and repository for natural language prompts}.

\bibitem[{Brown et~al.(2020{\natexlab{a}})Brown, Mann, Ryder, Subbiah, Kaplan,
  Dhariwal, Neelakantan, Shyam, Sastry, Askell, Agarwal, Herbert-Voss, Krueger,
  Henighan, Child, Ramesh, Ziegler, Wu, Winter, Hesse, Chen, Sigler, Litwin,
  Gray, Chess, Clark, Berner, McCandlish, Radford, Sutskever, and
  Amodei}]{NEURIPS2020_1457c0d6}
Tom Brown, Benjamin Mann, Nick Ryder, Melanie Subbiah, Jared~D Kaplan, Prafulla
  Dhariwal, Arvind Neelakantan, Pranav Shyam, Girish Sastry, Amanda Askell,
  Sandhini Agarwal, Ariel Herbert-Voss, Gretchen Krueger, Tom Henighan, Rewon
  Child, Aditya Ramesh, Daniel Ziegler, Jeffrey Wu, Clemens Winter, Chris
  Hesse, Mark Chen, Eric Sigler, Mateusz Litwin, Scott Gray, Benjamin Chess,
  Jack Clark, Christopher Berner, Sam McCandlish, Alec Radford, Ilya Sutskever,
  and Dario Amodei. 2020{\natexlab{a}}.
\newblock \href
  {https://proceedings.neurips.cc/paper_files/paper/2020/file/1457c0d6bfcb4967418bfb8ac142f64a-Paper.pdf}
  {Language models are few-shot learners}.
\newblock In \emph{Advances in Neural Information Processing Systems},
  volume~33, pages 1877--1901. Curran Associates, Inc.

\bibitem[{Brown et~al.(2020{\natexlab{b}})Brown, Mann, Ryder, Subbiah, Kaplan,
  Dhariwal, Neelakantan, Shyam, Sastry, Askell et~al.}]{brown2020language}
Tom Brown, Benjamin Mann, Nick Ryder, Melanie Subbiah, Jared~D Kaplan, Prafulla
  Dhariwal, Arvind Neelakantan, Pranav Shyam, Girish Sastry, Amanda Askell,
  et~al. 2020{\natexlab{b}}.
\newblock Language models are few-shot learners.
\newblock \emph{Advances in neural information processing systems},
  33:1877--1901.

\bibitem[{Chen and Shuai(2021)}]{chen2021meta}
Yi-Syuan Chen and Hong-Han Shuai. 2021.
\newblock Meta-transfer learning for low-resource abstractive summarization.
\newblock In \emph{Proceedings of the AAAI Conference on Artificial
  Intelligence}, volume~35, pages 12692--12700.

\bibitem[{Chen et~al.(2022)Chen, Liu, Xu, Yang, Zhu, Zeng, and
  Zhang}]{chen2022unisumm}
Yulong Chen, Yang Liu, Ruochen Xu, Ziyi Yang, Chenguang Zhu, Michael Zeng, and
  Yue Zhang. 2022.
\newblock Unisumm: Unified few-shot summarization with multi-task pre-training
  and prefix-tuning.
\newblock \emph{arXiv preprint arXiv:2211.09783}.

\bibitem[{Chowdhery et~al.(2022)Chowdhery, Narang, Devlin, Bosma, Mishra,
  Roberts, Barham, Chung, Sutton, Gehrmann et~al.}]{chowdhery2022palm}
Aakanksha Chowdhery, Sharan Narang, Jacob Devlin, Maarten Bosma, Gaurav Mishra,
  Adam Roberts, Paul Barham, Hyung~Won Chung, Charles Sutton, Sebastian
  Gehrmann, et~al. 2022.
\newblock Palm: Scaling language modeling with pathways.
\newblock \emph{arXiv preprint arXiv:2204.02311}.

\bibitem[{Cohan et~al.(2018)Cohan, Dernoncourt, Kim, Bui, Kim, Chang, and
  Goharian}]{cohan-etal-2018-discourse}
Arman Cohan, Franck Dernoncourt, Doo~Soon Kim, Trung Bui, Seokhwan Kim, Walter
  Chang, and Nazli Goharian. 2018.
\newblock \href {https://doi.org/10.18653/v1/N18-2097} {A discourse-aware
  attention model for abstractive summarization of long documents}.
\newblock In \emph{Proceedings of the 2018 Conference of the North {A}merican
  Chapter of the Association for Computational Linguistics: Human Language
  Technologies, Volume 2 (Short Papers)}, pages 615--621, New Orleans,
  Louisiana. Association for Computational Linguistics.

\bibitem[{Fabbri et~al.(2019)Fabbri, Li, She, Li, and
  Radev}]{fabbri-etal-2019-multi}
Alexander Fabbri, Irene Li, Tianwei She, Suyi Li, and Dragomir Radev. 2019.
\newblock \href {https://doi.org/10.18653/v1/P19-1102} {Multi-news: A
  large-scale multi-document summarization dataset and abstractive hierarchical
  model}.
\newblock In \emph{Proceedings of the 57th Annual Meeting of the Association
  for Computational Linguistics}, pages 1074--1084, Florence, Italy.
  Association for Computational Linguistics.

\bibitem[{Gao et~al.(2020)Gao, Biderman, Black, Golding, Hoppe, Foster, Phang,
  He, Thite, Nabeshima, Presser, and Leahy}]{pile}
Leo Gao, Stella Biderman, Sid Black, Laurence Golding, Travis Hoppe, Charles
  Foster, Jason Phang, Horace He, Anish Thite, Noa Nabeshima, Shawn Presser,
  and Connor Leahy. 2020.
\newblock The {P}ile: An 800gb dataset of diverse text for language modeling.
\newblock \emph{arXiv preprint arXiv:2101.00027}.

\bibitem[{Gliwa et~al.(2019)Gliwa, Mochol, Biesek, and
  Wawer}]{gliwa-etal-2019-samsum}
Bogdan Gliwa, Iwona Mochol, Maciej Biesek, and Aleksander Wawer. 2019.
\newblock \href {https://doi.org/10.18653/v1/D19-5409} {{SAMS}um corpus: A
  human-annotated dialogue dataset for abstractive summarization}.
\newblock In \emph{Proceedings of the 2nd Workshop on New Frontiers in
  Summarization}, pages 70--79, Hong Kong, China. Association for Computational
  Linguistics.

\bibitem[{Goyal et~al.(2022)Goyal, Li, and Durrett}]{goyal2022news}
Tanya Goyal, Junyi~Jessy Li, and Greg Durrett. 2022.
\newblock News summarization and evaluation in the era of gpt-3.
\newblock \emph{arXiv preprint arXiv:2209.12356}.

\bibitem[{He et~al.(2022)He, Peng, Lu, Wang, Mei, Liu, Xu, Awadalla, Shi, Zhu
  et~al.}]{he2022z}
Pengcheng He, Baolin Peng, Liyang Lu, Song Wang, Jie Mei, Yang Liu, Ruochen Xu,
  Hany~Hassan Awadalla, Yu~Shi, Chenguang Zhu, et~al. 2022.
\newblock Z-code++: A pre-trained language model optimized for abstractive
  summarization.
\newblock \emph{arXiv preprint arXiv:2208.09770}.

\bibitem[{Kim et~al.(2019)Kim, Kim, and Kim}]{kim-etal-2019-abstractive}
Byeongchang Kim, Hyunwoo Kim, and Gunhee Kim. 2019.
\newblock \href {https://doi.org/10.18653/v1/N19-1260} {Abstractive
  summarization of {R}eddit posts with multi-level memory networks}.
\newblock In \emph{Proceedings of the 2019 Conference of the North {A}merican
  Chapter of the Association for Computational Linguistics: Human Language
  Technologies, Volume 1 (Long and Short Papers)}, pages 2519--2531,
  Minneapolis, Minnesota. Association for Computational Linguistics.

\bibitem[{Koupaee and Wang(2018)}]{koupaee2018wikihow}
Mahnaz Koupaee and William~Yang Wang. 2018.
\newblock Wikihow: A large scale text summarization dataset.
\newblock \emph{arXiv preprint arXiv:1810.09305}.

\bibitem[{Lewis et~al.(2019)Lewis, Liu, Goyal, Ghazvininejad, Mohamed, Levy,
  Stoyanov, and Zettlemoyer}]{lewis2019bart}
Mike Lewis, Yinhan Liu, Naman Goyal, Marjan Ghazvininejad, Abdelrahman Mohamed,
  Omer Levy, Ves Stoyanov, and Luke Zettlemoyer. 2019.
\newblock Bart: Denoising sequence-to-sequence pre-training for natural
  language generation, translation, and comprehension.
\newblock \emph{arXiv preprint arXiv:1910.13461}.

\bibitem[{Li and Liang(2021)}]{li2021prefix}
Xiang~Lisa Li and Percy Liang. 2021.
\newblock Prefix-tuning: Optimizing continuous prompts for generation.
\newblock \emph{arXiv preprint arXiv:2101.00190}.

\bibitem[{Liang et~al.(2022)Liang, Bommasani, Lee, Tsipras, Soylu, Yasunaga,
  Zhang, Narayanan, Wu, Kumar et~al.}]{liang2022holistic}
Percy Liang, Rishi Bommasani, Tony Lee, Dimitris Tsipras, Dilara Soylu,
  Michihiro Yasunaga, Yian Zhang, Deepak Narayanan, Yuhuai Wu, Ananya Kumar,
  et~al. 2022.
\newblock Holistic evaluation of language models.
\newblock \emph{arXiv preprint arXiv:2211.09110}.

\bibitem[{Nenkova and McKeown(2012)}]{nenkova2012survey}
Ani Nenkova and Kathleen McKeown. 2012.
\newblock A survey of text summarization techniques.
\newblock \emph{Mining text data}, pages 43--76.

\bibitem[{Peng et~al.(2023)Peng, Li, He, Galley, and Gao}]{peng2023instruction}
Baolin Peng, Chunyuan Li, Pengcheng He, Michel Galley, and Jianfeng Gao. 2023.
\newblock Instruction tuning with gpt-4.
\newblock \emph{arXiv preprint arXiv:2304.03277}.

\bibitem[{Perez-Beltrachini and
  Lapata(2021)}]{perez-beltrachini-lapata-2021-models}
Laura Perez-Beltrachini and Mirella Lapata. 2021.
\newblock \href {https://doi.org/10.18653/v1/2021.emnlp-main.742} {Models and
  datasets for cross-lingual summarisation}.
\newblock In \emph{Proceedings of the 2021 Conference on Empirical Methods in
  Natural Language Processing}, pages 9408--9423, Online and Punta Cana,
  Dominican Republic. Association for Computational Linguistics.

\bibitem[{See et~al.(2017)See, Liu, and Manning}]{see-etal-2017-get}
Abigail See, Peter~J. Liu, and Christopher~D. Manning. 2017.
\newblock \href {https://doi.org/10.18653/v1/P17-1099} {Get to the point:
  Summarization with pointer-generator networks}.
\newblock In \emph{Proceedings of the 55th Annual Meeting of the Association
  for Computational Linguistics (Volume 1: Long Papers)}, pages 1073--1083,
  Vancouver, Canada. Association for Computational Linguistics.

\bibitem[{Sharma et~al.(2019)Sharma, Li, and Wang}]{sharma-etal-2019-bigpatent}
Eva Sharma, Chen Li, and Lu~Wang. 2019.
\newblock \href {https://doi.org/10.18653/v1/P19-1212} {{BIGPATENT}: A
  large-scale dataset for abstractive and coherent summarization}.
\newblock In \emph{Proceedings of the 57th Annual Meeting of the Association
  for Computational Linguistics}, pages 2204--2213, Florence, Italy.
  Association for Computational Linguistics.

\bibitem[{Smith et~al.(2022)Smith, Patwary, Norick, LeGresley, Rajbhandari,
  Casper, Liu, Prabhumoye, Zerveas, Korthikanti et~al.}]{smith2022using}
Shaden Smith, Mostofa Patwary, Brandon Norick, Patrick LeGresley, Samyam
  Rajbhandari, Jared Casper, Zhun Liu, Shrimai Prabhumoye, George Zerveas,
  Vijay Korthikanti, et~al. 2022.
\newblock Using deepspeed and megatron to train megatron-turing nlg 530b, a
  large-scale generative language model.
\newblock \emph{arXiv preprint arXiv:2201.11990}.

\bibitem[{Touvron et~al.(2023)Touvron, Lavril, Izacard, Martinet, Lachaux,
  Lacroix, Rozi{\`e}re, Goyal, Hambro, Azhar et~al.}]{touvron2023llama}
Hugo Touvron, Thibaut Lavril, Gautier Izacard, Xavier Martinet, Marie-Anne
  Lachaux, Timoth{\'e}e Lacroix, Baptiste Rozi{\`e}re, Naman Goyal, Eric
  Hambro, Faisal Azhar, et~al. 2023.
\newblock Llama: Open and efficient foundation language models.
\newblock \emph{arXiv preprint arXiv:2302.13971}.

\bibitem[{Wang et~al.(2022)Wang, Kordi, Mishra, Liu, Smith, Khashabi, and
  Hajishirzi}]{wang2022self}
Yizhong Wang, Yeganeh Kordi, Swaroop Mishra, Alisa Liu, Noah~A Smith, Daniel
  Khashabi, and Hannaneh Hajishirzi. 2022.
\newblock Self-instruct: Aligning language model with self generated
  instructions.
\newblock \emph{arXiv preprint arXiv:2212.10560}.

\bibitem[{Wei et~al.(2021)Wei, Bosma, Zhao, Guu, Yu, Lester, Du, Dai, and
  Le}]{wei2021finetuned}
Jason Wei, Maarten Bosma, Vincent~Y Zhao, Kelvin Guu, Adams~Wei Yu, Brian
  Lester, Nan Du, Andrew~M Dai, and Quoc~V Le. 2021.
\newblock Finetuned language models are zero-shot learners.
\newblock \emph{arXiv preprint arXiv:2109.01652}.

\bibitem[{Zhang et~al.(2020)Zhang, Zhao, Saleh, and Liu}]{zhang2020pegasus}
Jingqing Zhang, Yao Zhao, Mohammad Saleh, and Peter Liu. 2020.
\newblock Pegasus: Pre-training with extracted gap-sentences for abstractive
  summarization.
\newblock In \emph{International Conference on Machine Learning}, pages
  11328--11339. PMLR.

\bibitem[{Zhu and Zeng(2022)}]{impossibletriangle}
Chenguang Zhu and Michael Zeng. 2022.
\newblock Impossible triangle: What's next for pre-trained language models?
\newblock \emph{arXiv preprint arXiv:2204.06130}.

\end{thebibliography}
	\bibliographystyle{acl_natbib}
	
	\clearpage
 \onecolumn
	\appendix
	
	\section{List of Prompts \label{app:prompts}}
Below we list the prompts that we use from PromptSource. \texttt{[doc]} stands for the input document.\\
\texttt{[doc]  ===  Write a summary of the text above :  Summary:}\\
\texttt{[doc] How would you rephrase that in a few words?  Rephrase:}\\
\texttt{My college roommate asked me what this article means:  [doc]  So I recapped it in layman's terms:}\\
\texttt{Summarize this document: [doc]   Summary:}\\
\texttt{[doc]  ===  Given the above document, write one sentence to summarize:  Summary:}\\
\texttt{First, please read the article below.  [doc]  Now, can you write me an extremely short abstract for it?   An extremely short abstract:}\\
\texttt{[doc]  TL;DR:}\\
\texttt{Can you write an outline of the following article in a few points?  Article: [doc]  Outline:}\\
\texttt{Summarise the article:  [doc]  Summary:}\\
\texttt{In 2 or 3 sentences, what are the main points one should remember from this news article?  Article: [doc]  Main points:}\\
\texttt{Could you please generate a TLDR (Too Long Didn't Read) summary of the following news article?  Article: [doc] TLDR summary: }\\
\texttt{Condense the article down to the essentials to present it in the form of short cards in mobile news apps:  [doc]  Essentials:}\\
\texttt{Sum the following article in brief: [doc]  Breifs:}\\
\texttt{Extract key points from the article based on which the stock market could react:  [doc]  Key points:}\\
\texttt{Summarize this document: [doc]  Summary:}\\
\texttt{[doc] Given the above document, write a summary.  Summary:}\\
\texttt{Summarize: [doc]  Summary:}\\
\texttt{[doc] To sum up this document:}\\
\texttt{Sum up the following document:  [doc]  Summary:}\\
\texttt{What are the key points across these news articles:  Article: [doc]  Key points:}\\
\texttt{Synthesize these documents into a single one:  - [doc]  Summary:}\\
\texttt{I want to edit the following articles into a more concise summary:  Article: [doc]  Summary:}\\
\texttt{Write a summary of the following articles:  Document: [doc]  Summary:}\\
\texttt{I'm trying to distill these articles down into one:  Article: [doc]  Summary:}

The special ``following'' prompt is\\
\texttt{"Follow the example(s) above and summarize the document below: Document: [doc] Summary:}\\
The in-context examples are prepended to the prompt in the format of \texttt{Document: [doc] Summary: [sum]} where \texttt{[doc]} and \texttt{[sum]} are the document and summary for the in-context examples respectively.

	
		
 \section{Sampling ratio of tasks}
We mix \Circled{1} and \Circled{2} as one task, and treat \Circled{3} \Circled{4}	\Circled{5} \Circled{6} as individual tasks. They are mixed by the ratio of [0.45,0.1,0.15,0.15, 0.15].
\end{document}